\documentclass[letterpaper]{article} % DO NOT CHANGE THIS
\usepackage{aaai2026}  % DO NOT CHANGE THIS
\usepackage{times}  % DO NOT CHANGE THIS
\usepackage{helvet}  % DO NOT CHANGE THIS
\usepackage{courier}  % DO NOT CHANGE THIS
\usepackage[hyphens]{url}  % DO NOT CHANGE THIS
\usepackage{graphicx} % DO NOT CHANGE THIS
\usepackage{natbib}  % DO NOT CHANGE THIS AND DO NOT ADD ANY OPTIONS TO IT
\usepackage{caption} % DO NOT CHANGE THIS AND DO NOT ADD ANY OPTIONS TO IT
\usepackage{algorithm}
\usepackage{algorithmic}
\usepackage{amsfonts}       % blackboard math symbols
\usepackage{amsmath}
\usepackage{booktabs}       % professional-quality tables
\usepackage{multirow}
\usepackage{newfloat}
\usepackage{listings}
\DeclareCaptionStyle{ruled}{labelfont=normalfont,labelsep=colon,strut=off} % DO NOT CHANGE THIS
\floatstyle{ruled}
\newfloat{listing}{tb}{lst}{}
\floatname{listing}{Listing}
\nocopyright 
\title{2D Gaussians Meet Visual Tokenizer}

\author{
    \textnormal{Yiang Shi\textsuperscript{1,2}},
    \textnormal{Xiaoyang Guo\textsuperscript{2}\thanks{Project lead}},
    \textnormal{Wei Yin\textsuperscript{2}},
    \textnormal{Mingkai Jia\textsuperscript{2}}, \\
    \textnormal{Qian Zhang\textsuperscript{2}},
    \textnormal{Xiaolin Hu\textsuperscript{3}},
    \textnormal{Wenyu Liu\textsuperscript{1}},
    \textnormal{Xinggang Wang\textsuperscript{1}\thanks{Corresponding author: xgwang@hust.edu.cn}} \\
    \textsuperscript{1}Huazhong University of Science and Technology \\
    \textsuperscript{2}Horizon Robotics \\
    \textsuperscript{3}Department of Computer Science and Technology, Tsinghua University
}

\newcommand{\name}{VGQ}
\begin{document}

\maketitle

\begin{abstract}
The image tokenizer is a critical component in AR image generation, as it determines how rich and structured visual content is encoded into compact representations. Existing quantization-based tokenizers such as VQ-GAN~\cite{vqgan} primarily focus on appearance features like texture and color, often neglecting geometric structures due to their patch-based design. In this work, we explored how to incorporate more visual information into the tokenizer and proposed a new framework named Visual Gaussian Quantization (VGQ), a novel tokenizer paradigm that explicitly enhances structural modeling by integrating 2D Gaussians into traditional visual codebook quantization frameworks. Our approach addresses the inherent limitations of naive quantization methods such as VQ-GAN, which struggle to model structured visual information due to their patch-based design and emphasis on texture and color. In contrast, VGQ encodes image latents as 2D Gaussian distributions, effectively capturing geometric and spatial structures by directly modeling structure-related parameters such as position, rotation and scale.
We further demonstrate that increasing the density of 2D Gaussians within the tokens leads to significant gains in reconstruction fidelity, providing a flexible trade-off between token efficiency and visual richness.
On the ImageNet $256 \times 256$ benchmark, VGQ achieves strong reconstruction quality with an rFID score of 1.00. Furthermore, by increasing the density of 2D Gaussians within the tokens, VGQ gains a significant boost in reconstruction capability and achieves a \textbf{state-of-the-art reconstruction rFID score of 0.556} and a PSNR of 24.93, substantially outperforming existing methods. Codes will be released soon.
\end{abstract}

\section{Introduction}

Image generation has witnessed rapid advancements in recent years~\cite{magvit1,magvit2,llamagen,vqvae}, drawing growing interest from both academia and industry. With the success of large language models (LLMs)~\cite{llama2, palm,llava,openai2023gpt4,gpt3,minigpt}, discrete image generation has emerged as a promising direction, as it allows the adoption of autoregressive modeling paradigms originally developed for natural language processing. Central to these approaches is the image tokenizer \cite{open-magvit2,titok,tokenflow,flextok}, which transforms raw pixel data into discrete latents. The tokenizer not only facilitates image compression but also supports cross-modal integration and scalable model training. Therefore, designing an effective tokenizer that retains rich visual information is critical for high-quality image generation.

In this paper, we first ask ourselves: \textit{how can we empower the tokenizer to acquire more visual information?} To address this issue, we reflect on the existing popular tokenizers. Most of the popular tokenizers today are based on VQ-GAN \cite{vqgan}, which discretizes images by patchifying them and encoding each patch into a codebook index. While effective, this approach tends to overlook the intrinsic geometric and structural properties of images. The patchify operations often blur spatial boundaries and compromise fine-grained spatial relationships, resulting in suboptimal tokenization. A typical challenge is that tokenizers usually struggle to reconstruct images containing text, as even slight distortion of structural information can render the text completely unrecognizable. Then, we drew inspiration from the rapidly evolving Gaussian Splatting technique, believing its recent advancements hold great potential for designing new tokenizer architectures.

2D Gaussian Splatting \cite{2dgs}, as a method originally proposed for 3D reconstruction, has recently been extended to image representation \cite{gsimg,image-gs,gstk} due to its strong capability in modeling structural information. It represents the image into \textbf{learnable ellipses of varying shapes and sizes}, which, unlike simple grid-based partitioning, enables the model to capture structured information within the image. While patch-based tokenization (such as VQ-GAN) relies on dividing the image into uniform patches and treats each patch as an isolated feature, 2D Gaussians maintain continuous, flexible representations of the image’s spatial structure. By introducing 2D Gaussians into the tokenizer, \name{} is capable of capturing information about the relative positioning and orientation of objects in the image. To further exploit the structural modeling capability of 2D Gaussians, we introduce an extended version of our tokenizer, named \textbf{\name{}-multigs}, which increases the density of 2D Gaussians within tokens. This design significantly increases the representational capacity of the model, enabling it to better capture complex geometric and spatial relationships. Our experiments in Table \ref{tab:main} and visualization in Fig. \ref{fig:vis} show that \name{}-multigs achieves substantial improvements in reconstruction fidelity, especially in areas containing fine-grained structures such as text and edges, demonstrating the effectiveness of multi-Gaussian encoding.

To sum up, in this paper, we introduce Visual Gaussian Quantization, a novel tokenizer architecture that incorporates 2D Gaussian Splatting into the tokenization pipeline. Specifically, we design a dual-branch architecture: one branch follows the conventional VQ-GAN path, while the other encodes latents using 2D Gaussians. For the branch using 2D Gaussians, we define discrete codebooks over both the Gaussian parameters (e.g., scale, rotation, and position) and the associated features. During detokenization, fully discretized 2D Gaussians are splatted to reconstruct the feature map, which is then fused with the output from the VQ branch. By leveraging the structural modeling capacity of 2D Gaussians, our tokenizer can capture fine-grained geometry and layout, leading to significant improvements in image reconstruction and generation quality.

Our main contributions are summarized as follows:
\begin{itemize}
    \item We propose the first tokenizer architecture that integrates fully discretized 2D Gaussians into the image tokenization process named \name{} and \name{} achieves strong performance with an rFID of 1.00.
    \item We demonstrate that incorporating 2D Gaussians enables the preservation of fine details and geometric structures that are often lost in conventional VQ-based approaches.
    \item We extend our model to \name{}-multigs, allowing each token to encode multiple 2D Gaussians, thereby enhancing the representation capacity for complex geometric structures.
    \item \name{}-multigs achieves \textbf{state-of-the-art} reconstruction performance, with an rFID score of 0.556 and a PSNR of 24.93 on the ImageNet dataset, significantly outperforming existing methods.
\end{itemize}

\section{Related Work}

\subsection{Image Tokenizer}

Image tokenizer plays a foundational role in discrete image generation by transforming continuous image data into a sequence of discrete tokens. Formally, given an image $x \in \mathbb{R}^{H \times W \times C}$, where $H$, $W$, and $C$ denote the height, width, and number of channels, respectively, a tokenizer $f_\theta$ maps the image into a sequence of tokens $z = f_\theta(x) = [z_1, z_2, \ldots, z_N]$, where each $z_i \in \mathcal{V}$ and $\mathcal{V}$ is a discrete vocabulary of size $K$. One of the most widely adopted approaches is the \textit{Vector Quantization} (VQ) technique introduced in VQ-GAN~\cite{vqgan}. In this framework, the tokenizer is trained to minimize the reconstruction loss while enforcing each embedding $e_i$ to be selected from a learned codebook $\mathcal{E} = \{e_1, e_2, \ldots, e_K\} \subset \mathbb{R}^d$:
\begin{equation}
    \mathcal{L}_{\text{VQ}} = \|x - \hat{x}\|^2 + \| \text{sg}[z_e(x)] - e_{q(x)} \|^2 + \beta \| z_e(x) - \text{sg}[e_{q(x)}] \|^2,
\end{equation}
where $\text{sg}[\cdot]$ denotes the stop-gradient operator, $z_e(x)$ is the encoder output, and $e_{q(x)}$ is the nearest codebook entry.
100,000
Building upon VQ-GAN, numerous works have proposed improvements to the image tokenizer. FSQ~\cite{fsq} replaces vector quantization in VQ-VAEs with a simple and effective scalar quantization scheme, achieving high codebook utilization and competitive performance across image generation and vision tasks without requiring auxiliary losses or codebook tricks. VQGAN-LC~\cite{vqganlc} uses a frozen codebook initialized from pre-trained vision features and trains a lightweight projector, achieving 99\% utilization on a large codebook with size 100,000. RQ-VAE~\cite{rqvae} and DQ-VAE~\cite{dqvae} enhance VQ-GAN by incorporating residual quantization and dynamic quantization techniques, respectively. MAGVIT-v2\cite{magvit2} introduces a unified visual tokenizer for both images and videos, leveraging a novel lookup-free quantization (LFQ) mechanism to expand the codebook size. TiTok~\cite{titok} proposes a transformer-based 1D image tokenizer that compresses a 256×256 image into as few as 32 discrete tokens. FlexTok~\cite{flextok} proposes a flexible 1D tokenization scheme using nested dropout and rectified flow decoding, allowing autoregressive models to generate images with a variable number of tokens based on semantic complexity.

\subsection{Gaussian Splatting}
3D Gaussian Splatting (3DGS)~\cite{3dgs} introduces a novel method for representing 3D scenes using Gaussian primitives, which are rendered into 2D images via differentiable volumetric splatting. Building upon this foundation, numerous reconstruction methods~\cite{4dgs, halfgs} have been developed to further enhance efficiency and fidelity.
 Each Gaussian primitive is explicitly parameterized by its 3D location $\mathbf{p_k}$ and a covariance matrix $\boldsymbol{\Sigma}$, defining the spatial distribution of its density. The density function of a 3D Gaussian centered at $\mathbf{p_k}$ is given by:
\begin{equation}
\mathcal{G}(\mathbf{p}) = \exp\left(-\frac{1}{2} (\mathbf{p} - \mathbf{p}_k)^\top \boldsymbol{\Sigma}^{-1} (\mathbf{p} - \mathbf{p}_k)\right)
\label{eq:gaussian}
\end{equation}

However, 3DGS often suffers from geometric inconsistency due to the volumetric nature of its primitives. To address this, Huang et al. propose \textbf{2D Gaussian Splatting (2DGS)}~\cite{2dgs}, which collapses 3D Gaussians into oriented 2D elliptical disks embedded in 3D space. Each 2D Gaussian is parameterized by a center $\mathbf{p}_k$, tangential vectors $\mathbf{t}_u, \mathbf{t}_v$, and scales $s_u, s_v$, forming a local geometry matrix:
\begin{equation}
H = \begin{bmatrix}
\mathbf{t}_u s_u & \mathbf{t}_v s_v & \mathbf{t}_w \cdot 0 & \mathbf{p}_k \\
0 & 0 & 0 & 1
\end{bmatrix},
\end{equation}

where $\mathbf{t}_w = \mathbf{t}_u \times \mathbf{t}_v$ defines the surface normal.
2DGS enables perspective-accurate ray-splat intersection and real-time rendering via alpha-blending:
\begin{equation}
c(\mathbf{x}) = \sum_{i} c_i \alpha_i \hat{G}_i(\mathbf{u}(\mathbf{x})) \prod_{j < i} (1 - \alpha_j \hat{G}_j(\mathbf{u}(\mathbf{x}))),
\end{equation}
where $\hat{G}_i$ denotes the regularized Gaussian response at ray-splat intersection point $\mathbf{u}(\mathbf{x})$.

To improve reconstruction quality, 2DGS introduces two regularization losses: \textit{depth distortion} $\mathcal{L}_d$ and \textit{normal consistency} $\mathcal{L}_ns$, optimizing the final objective:
\begin{equation}
\mathcal{L} = \mathcal{L}_c + \alpha \mathcal{L}_d + \beta \mathcal{L}_n,
\end{equation}
where $\mathcal{L}_c$ is a photometric loss. Experimental results show that 2DGS achieves great performance in surface reconstruction with significantly improved geometric fidelity and rendering efficiency compared to 3DGS and implicit methods.

\section{Method}

This section introduces our proposed Visual Gaussian Quantization that jointly leverages vector quantization and 2D Gaussian Splatting. Our goal is to build a compact yet expressive image representation by capturing both local appearance and global geometric structure. To this end, we design a dual-branch tokenization module, where each branch encodes complementary aspects of the input image. The final representation is fused and decoded for reconstruction, and the entire model is trained in an end-to-end fashion.

\subsection{Overview}

Our framework consists of a shared encoder, two parallel token branches (VQ and 2DGS), a fusion module, and a shared decoder. Given an input image $I \in \mathbb{R}^{H_0 \times W_0 \times 3}$, a convolutional encoder extracts a dense feature map $F \in \mathbb{R}^{H \times W \times d}$. This feature map is simultaneously processed by two complementary tokenization modules:

\begin{itemize}
    \item The \textbf{VQ branch} discretizes local appearance patterns using a vector-quantized codebook.
    \item The \textbf{2DGS branch} encodes structural information via learnable 2D Gaussians.
\end{itemize}

The two token sets are fused into a unified representation, which is then decoded into an RGB image. This architecture enables the model to leverage both photometric and structural inductive biases. Our training pipeline is shown in~\ref{algorithm}.

\begin{algorithm}[tb]
\caption{\name{} pipeline}
\begin{lstlisting}[language=Python]
def fusion(vq_tokens, gs_tokens):
    # Element-wise multiplication for feature fusion
    return vq_tokens * gs_tokens
def tokenize_2dgs(features, codebook_geo, codebook_feat):
    tokens = []
    for feat in features:
        geo_params, feat_params = feat[:5], feat[5:] 
        # geo_params : position,rotation,scale
        geo_dist = torch.norm(geo_params - codebook_geo, dim=1)
        feat_dist = torch.norm(feat_params - codebook_feat, dim=1)
        geo_idx, feat_idx = torch.argmin(geo_dist), torch.argmin(feat_dist)
        tokens.append((codebook_geo[geo_idx], codebook_feat[feat_idx]))
    return tokens
def train_step(model, data, codebook_geo, codebook_feat):
    # 1. dual-branch tokenization
    gs_tokens = tokenize_2dgs(data, codebook_geo, codebook_feat)
    vq_tokens = tokenize_vq(data)
    fused_tokens = fusion(vq_tokens, gs_tokens)
    # 2. reconstruction
    reconstructed = model.decode(fused_tokens)
    loss = (reconstructed - data).pow(2).mean()
    loss.backward()

\end{lstlisting}
\label{algorithm}
\end{algorithm}

\subsection{VQ Branch}

To capture local texture, we employ conventional vector quantization over the spatial feature grid. For each spatial location $(h,w)$ in the encoder output, the feature $\mathbf{f}_{h,w} \in \mathbb{R}^d$ is mapped to the nearest entry in a learned codebook $\mathcal{C}_\text{VQ} = \{\mathbf{e}_j\}_{j=1}^{K_\text{VQ}}$:

\begin{equation}
z_{h,w} = \arg\min_j \|\mathbf{f}_{h,w} - \mathbf{e}_j\|_2^2, \quad 
F_\text{VQ}(h,w) = \mathbf{e}_{z_{h,w}}.
\end{equation}

The codebook is updated using an exponential moving average (EMA) of encoder features assigned to each entry. While this branch effectively captures local appearance, it lacks explicit modeling of geometry.
 % following the original VQ-VAE~\cite{vqvae}
\subsection{2DGS Branch}

To complement the VQ branch, the 2DGS branch explicitly models image structure via a set of spatially localized 2D Gaussians. Each image is represented by $N=256$ discrete tokens, where each token corresponds to a 2D Gaussian $g_k = (\mathbf{p}_k, \theta_k, s_k, \mathbf{f}_k)$. Here, $\mathbf{p}_k \in \mathbb{R}^2$ denotes the 2D position on the image plane, $\theta_k \in [0, 2\pi)$ is the rotation angle, $s_k \in \mathbb{R}^2$ is the scale vector specifying the axes of the Gaussian, and $\mathbf{f}_k \in \mathbb{R}^d$ is a local feature embedding. All these parameters are learnable. We define two codebooks to discretize the parameters. Both quantizations are performed via nearest-neighbor matching:
\begin{equation}
\mathcal{C}_\text{geo} = \{(\hat{\mathbf{p}}_i, \hat{\theta}_i, \hat{s}_i)\}_{i=1}^{K_\text{geo}}, \quad 
\mathcal{C}_\text{feat} = \{\hat{\mathbf{f}}_j\}_{j=1}^{K_\text{feat}}.
\end{equation}

\begin{equation}
z_k^\text{geo} = \arg\min_i \|(\mathbf{p}_k, \theta_k, s_k) - (\hat{\mathbf{p}}_i, \hat{\theta}_i, \hat{s}_i)\|_2^2,
\end{equation}
\begin{equation}
z_k^\text{feat} = \arg\min_j \|\mathbf{f}_k - \hat{\mathbf{f}}_j\|_2^2.
\end{equation}

To improve alignment of these 2D Gaussians with the original spatial grid, we introduce a deformable cross-attention mechanism. The parameters of all tokens, including their 2DGS attributes (such as position, scale, and orientation), are updated through a deformable attention mechanism. The attention mechanism allows each token's 2D Gaussian to focus on adaptive regions of the encoder features, with its spatial parameters (e.g., position and scale) updated during training, leading to improved alignment and structurally grounded representations. All geometric parameters are constrained to valid ranges (e.g., positive scale, bounded angles), but no specialized initialization is required. In the detokenization phase, Gaussians are splatted into a continuous feature map:

\begin{equation}
F_\text{2DGS}(\mathbf{x}) = \sum_{k=1}^{N} \alpha_k \cdot \mathcal{G}_k(\mathbf{x}) \cdot \hat{\mathbf{f}}_{z_k^\text{feat}},
\end{equation}

where $\mathcal{G}_k(\mathbf{x})$ is a Gaussian kernel defined by $\mathcal{C}_\text{geo}$, and $\alpha_k$ is a learnable opacity scalar.

\subsection{Fusion with VQ Branch}

To merge the two representations, we adopt the Hadamard product (element-wise multiplication)  over the aligned feature maps:

\begin{equation}
F_\text{final}(h,w) = F_\text{VQ}(h,w) \odot F_\text{2DGS}(h,w).
\label{eq_fuse}
\end{equation}

This fusion method allows the structure-aware 2DGS features to act as spatial gates that modulate texture features learned by the VQ branch. As can be seen in the experimental section, this strategy performs better than the additive or attention mechanisms.

\subsection{Decoder and Reconstruction}

The fused feature map $F_\text{final}$ is passed through a convolutional decoder to generate the reconstructed image $\hat{I} \in \mathbb{R}^{H_0 \times W_0 \times 3}$. The decoder mirrors the encoder in structure and progressively upsamples the latent features to match the original image resolution.
We adopt adversarial training using a PatchGAN-style discriminator~\cite{patchgan}.
The training objective combines reconstruction loss, adversarial loss, and perceptual loss:
\begin{align}
\mathcal{L}_\text{rec} &= \|I - \hat{I}\|_1, \\
\mathcal{L}_\text{adv} &= \mathbb{E}[(D(\hat{I}) - 1)^2 + (D(I))^2], \\
\mathcal{L}_\text{perceptual} &= \sum_l \|\phi_l(I) - \phi_l(\hat{I})\|_2^2.
\end{align}
The reconstruction loss encourages pixel-level accuracy. The adversarial loss guides the generator toward producing more realistic outputs. The perceptual loss compares the high-level features of the generated image and the ground truth, extracted using a pretrained VGG network~\cite{vgg}. In the perceptual loss, $\phi_l(\cdot)$ represents the activation from the $l$-th layer of the VGG model.
The total training loss is: 
\begin{equation}
\mathcal{L} = \mathcal{L}_\text{rec} + \gamma \cdot \mathcal{L}_\text{adv} + \eta \cdot \mathcal{L}_\text{perceptual},
\end{equation}

\subsection{Codebook Update}

Both codebooks $\mathcal{C}_\text{VQ}$ and $\mathcal{C}_\text{feat}$/$\mathcal{C}_\text{geo}$ are updated using exponential moving average. Given an assigned feature $\mathbf{f}$ and codebook vector $\mathbf{e}_j$, the update rule is:

\begin{equation}
\mathbf{e}_j \leftarrow m \cdot \mathbf{e}_j + (1 - m) \cdot \mathbf{f},
\end{equation}

where $m$ is the EMA decay coefficient. This mechanism stabilizes the discrete latent space and avoids noisy fluctuations during training.

\section{Theoretical Analysis}

This section provides the theoretical foundation for our dual-branch tokenization approach. We begin by presenting a framework based on image content decomposition, followed by an explanation of how our proposed 2DGS component enhances conventional VQ tokenization, improving its ability to encode structural information.

\subsection{Information-Theoretic Formulation}

We model an image $I$ as originating from a generative process with two latent factors: an appearance factor $A$ and a structural factor $S$. This decomposition is a well-established approach, previously adopted by various methods such as \cite{disentangle-obj-appearance, img-decomposition}. This decomposition is expressed as:
\begin{equation}
I = f(A, S),
\label{eq:image_decomposition_ch}
\end{equation}
where $A$ represents local textural and chromatic features, while $S$ describes global geometry and perceptual grouping. The tokenization process $\mathcal{T}$ aims to preserve information from both factors. The fidelity of this process can be quantified by the mutual information:
\begin{equation}
\mathcal{J}(Z) = I(Z; A) + I(Z; S).
\label{eq:total_mi_ch}
\end{equation}
Maximizing $\mathcal{J}(Z)$ is the goal of an ideal tokenizer. Analyzing methods applying information theory is also a common approach, as demonstrated in works such as \cite{mutual-information}.

\subsection{Limitations of Standard VQ Tokenization}

Vector Quantization (VQ) methods focus on spatially independent quantization over image features. While effective for encoding appearance-related information (high $I(Z_\text{VQ}; A)$), they struggle to capture global structural factors like object localization and geometric relationships. The limitations stem from two key factors:
\begin{itemize}
    \item Token indices are mapped to a fixed grid, preventing precise encoding of spatial relationships.
    \item Basic geometric attributes such as scale and orientation are not explicitly modeled.
\end{itemize}
As a result, VQ tokenization is effective for appearance, but fails to encode complex spatial structures.
To address these limitations, we propose the 2D Gaussian Splatting (2DGS) component, which introduces a parametric encoding of geometric primitives. Each token in $Z_\text{2DGS}$ is represented by:
\begin{equation}
Z_\text{2DGS} = \{ (\mathbf{p}_k, \theta_k, s_k, \mathbf{f}_k) \}_{k=1}^N.
\label{eq:2dgs_token_ch}
\end{equation}
This representation includes position $\mathbf{p}_k$, orientation $\theta_k$, scale $s_k$, and feature $\mathbf{f}_k$. These parameters are linked to the structural factor $S$, allowing $Z_\text{2DGS}$ to capture geometric details and enhance $I(Z_\text{2DGS}; S)$.

\subsection{Synergistic Combination of $Z_\text{VQ}$ and $Z_\text{2DGS}$}

The combination of $Z_\text{VQ}$ and $Z_\text{2DGS}$ into a unified representation $Z_{\text{comb}} = Z_{\text{VQ}} \cup Z_{\text{2DGS}}$ leverages their complementary strengths. The goal is to maximize $\mathcal{J}(Z_{\text{comb}})$ by enhancing structural encoding.
Using the chain rule for mutual information, the structural information captured by $Z_{\text{comb}}$ is:
\begin{equation}
\begin{split}
I(Z_{\text{comb}}; S) = I(Z_{\text{VQ}}, Z_{\text{2DGS}}; S) = \\
I(Z_{\text{VQ}}; S) + I(Z_{\text{2DGS}}; S | Z_{\text{VQ}}).
\label{eq:combined_mi_S_ch}
\end{split}
\end{equation}
% \begin{align}
% I(Z_{\text{comb}}; S) = I(Z_{\text{VQ}}, Z_{\text{2DGS}}; S) = \\
% I(Z_{\text{VQ}}; S) + I(Z_{\text{2DGS}}; S | Z_{\text{VQ}}).
% \label{eq:combined_mi_S_ch}
% \end{align}
This shows that $I(Z_{\text{comb}}; S)$ includes the structural information from $Z_{\text{VQ}}$ and the additional structural information from $Z_{\text{2DGS}}$ conditioned on $Z_{\text{VQ}}$. The term $I(Z_{\text{2DGS}}; S | Z_{\text{VQ}})$ reflects the new structural information provided by $Z_{\text{2DGS}}$, which $Z_{\text{VQ}}$ does not capture.
For comparison, consider an alternative approach $Z'_{\text{comb}} = Z_{\text{VQ1}} \cup Z_{\text{VQ2}}$, where two VQ modules are independently optimized. The mutual information with $S$ is:
\begin{equation}
\begin{split}
I(Z'_{\text{comb}}; S) = I(Z_{\text{VQ1}}, Z_{\text{VQ2}}; S) \\
= I(Z_{\text{VQ1}}; S) + I(Z_{\text{VQ2}}; S | Z_{\text{VQ1}}).
\end{split}
\label{eq:dual_vq_mi_S_ch}
\end{equation}
Since $Z_{\text{VQ2}}$ is likely to capture similar structural information as $Z_{\text{VQ1}}$, the incremental gain $I(Z_{\text{VQ2}}; S | Z_{\text{VQ1}})$ is limited.
In contrast, $Z_{\text{2DGS}}$ is designed to capture geometric details that $Z_{\text{VQ}}$ misses. Therefore, the differential information gain from $Z_{\text{2DGS}}$ ($I(Z_{\text{2DGS}}; S | Z_{\text{VQ}})$) is significantly greater than the gain from a second VQ encoder, leading to the inequality:
\begin{equation}
I(Z_{\text{VQ1}}, Z_{\text{VQ2}}; S) < I(Z_{\text{VQ}}, Z_{\text{2DGS}}; S).
\label{eq:comparison_inequality_ch}
\end{equation}
This demonstrates that the $Z_{\text{VQ}} \cup Z_{\text{2DGS}}$ combination better preserves structural fidelity, due to the complementary nature of 2DGS.

\subsection{Conclusion}

In conclusion, the integration of the 2DGS module significantly improves the fidelity and expressiveness of the tokenizer. By enhancing both appearance features ($I(Z;A)$) and structural details ($I(Z;S)$), our approach maximizes the information-theoretic objective $\mathcal{J}(Z)$.
While VQ tokenization is effective for appearance encoding, its design limits its ability to capture structural information. The 2DGS module, however, provides a dedicated mechanism for encoding geometric context, improving the overall representation. Experiments shown in Table \ref{tab:mini_result} support our analysis.

\section{Experiments and Ablations}

To evaluate the effectiveness and generalizability of our proposed tokenizer \name{}, we conduct experiments on both Mini-ImageNet and ImageNet-1k datasets, using a resolution of $256 \times 256$. We first evaluate on Mini-ImageNet to validate the core design of our tokenizer. Based on these insights, we scale up training to the full ImageNet dataset and report \textbf{state-of-the-art results}. We report results using three key metrics: \textbf{rFID}, \textbf{PSNR}, and \textbf{codebook utilization rate}. 

\subsection{Validation on Mini-ImageNet}

We begin by testing \name{} on the Mini-ImageNet dataset.
Mini-ImageNet is a small benchmark derived from the ImageNet-1K dataset. This dataset contains 100 distinct classes, each comprising 600 images. Following standard practice, we use 48,000 images for training and the remaining 12,000 images for evaluation.
We aim to verify that incorporating 2DGS into the tokenizer leads to better performance than VQ. 
To make a fair comparison, we experimented with fusing two independently constructed VQ-based feature maps.
This fusion of two independent VQ branches performs worse than the fusion of VQ and 2DGS, as shown in Table~\ref{tab:mini_result}.
These findings further demonstrate the positive impact of incorporating 2DGS, which is consistent with the analysis presented in the previous section.

% \input{mini_result}
% \begin{table*}[t]
\begin{table*}[t]
    \centering
    \footnotesize
    \setlength{\tabcolsep}{4pt}
    \renewcommand\arraystretch{1.1}
    % 使用 minipage + \centering 强制居中
    \begin{minipage}{\textwidth}
    \centering
    \resizebox{\textwidth}{!}{ % 改为 \textwidth 更稳定
        \begin{tabular}{lccccccccc}
        \toprule
        \multirow{2}{*}{\textbf{Method}} & \textbf{Token} & \multirow{2}{*}{\textbf{Tokens}} & \multirow{2}{*}{\textbf{Ratio}} & \textbf{Train} & \textbf{Codebook} & \textbf{Codebook} & \multirow{2}{*}{\textbf{rFID}$\downarrow$} & \multirow{2}{*}{\textbf{PSNR}$\uparrow$} & \multirow{2}{*}{\textbf{SSIM}$\uparrow$} \\ 
        & \textbf{Type} &  & & \textbf{Resolution} & \textbf{Size} & \textbf{Dim}& & & \\
        \midrule
        VQGAN~\cite{vqgan} & 2D & 16 $\times$ 16 & 16 & 256 $\times$ 256 & 1,024  & 256 & 32.31 & 18.71 & 0.50 \\
        VQGAN$\star$ & 2D & 16 $\times$ 16 & 16 & 256 $\times$ 256 & 1,024  & 256 & 22.81 & 19.37 & 0.53 \\
        \name{} & 2D & 16 $\times$ 16 & 16 & 256 $\times$ 256 & 1,024  & 256 & \textbf{13.11} & \textbf{20.08} & \textbf{0.57} \\
        % IBQ & 2D & 16 $\times$ 16 & 16 & 256 $\times$ 256 & 16,384 & 256 & $1.37$ & 0.2235 & 96$\%$ \\
        % IBQ & 2D & 16 $\times$ 16 & 16 & 256 $\times$ 256 & 262,144 & 256 & 1.00 & 0.2030 & 84\% \\
        % \textbf{GQT} & 2D & 16 $\times$ 16 & 16 &256 $\times$ 256 & 16384 & 8 & \textbf{1.00}  &22.91 & \textbf{0.1639} & 100\% \\
        % \hline
        % Titok-L~\citep{titok} & 1D & 32 & $-$ & $256 \times 256$ & 4,096 & 16 & 2.21 & $-$ & $-$  & $-$ \\
        % Titok-B~\citep{titok} & 1D & 64 & $-$ & $256 \times 256$ & 4,096 & 16& 1.70 & $-$ & $-$  & $-$\\
        % Titok-S~\citep{titok} & 1D & 128 & $-$ & $256 \times 256$ & 4,096 & 16 & 1.71 & $-$ & $-$ & $-$ \\        
        \bottomrule
        \end{tabular}
    }
    \caption{Reconstruction performance of different tokenizers on $256 \times 256$ Mini-ImageNet. $\star$ means that the features are obtained by fusing two independently trained VQ branches.}
    \label{tab:mini_result}
    \end{minipage}
\end{table*}

\subsection{Ablation on Mini-ImageNet}

A natural extension of our design is to allow each token to contain more than one 2D Gaussian. Intuitively, when a token encodes multiple Gaussians, it can carry richer structural information. To explore this idea, we conducted experiments on Mini-ImageNet by increasing the number of Gaussians per token. 
Results shown in Table \ref{tab:num_gs} indicate an encouraging observation that reconstruction quality improves consistently as the number of Gaussians increases, indicating that our 2DGS-based representation scales well with increased structural capacity.

Furthermore, we conducted an ablation study on the feature fusion strategy and compared three different approaches. \textit{Cross Attention} refers to computing attention between the VQ and 2DGS feature maps. \textit{Mask Adding} denotes a learnable mechanism that predicts a spatial mask over the VQ features, which is then used to modulate the VQ features before adding them to the 2DGS features. \textit{Hadamard Product}, our final choice, performs element-wise multiplication between the VQ and 2DGS feature maps, allowing structural features to act as soft gates for modulating appearance information.  Results shown in Table \ref{tab:fusion} prove that Hadamard Product outperforms both of the alternative fusion strategies in terms of reconstruction quality.

% \textit{Cross Attention} refers to computing attention between the VQ and 2DGS feature maps. \textit{Mask Add} denotes a learnable mechanism that predicts a spatial mask over the VQ features, which is then used to modulate the VQ features before adding them to the 2DGS features. Experimental results show that \textbf{Hadamard Product} outperforms both of these fusion strategies in terms of reconstruction quality.

% 表格代码
% \begin{figure}[t]
\begin{figure}[t]
\centering
\begin{minipage}{0.48\textwidth}
    \centering
    \resizebox{\textwidth}{!}{
        \begin{tabular}{c|ccc}
            \toprule
            \ Fusion Method & rFID$\downarrow$ & PSNR$\uparrow$  & SSIM$\uparrow$ \\
            \midrule
            Cross Attention  & 34.22 & 18.87 & 0.50 \\
            Mask Adding  & 13.46 & 19.72 & \textbf{0.57} \\
            Hadamard Product & \textbf{13.11} & \textbf{20.08} & \textbf{0.57} \\
            \bottomrule
        \end{tabular}
    }
    \captionof{table}{\textbf{Ablation of Fusion Methods.} Hadamard Product achieves the best performance among all fusion strategies.}
    \label{tab:fusion}
\end{minipage}
\hfill
\begin{minipage}{0.48\textwidth}
    \centering
    \resizebox{\textwidth}{!}{
        \begin{tabular}{c|ccc}
            \toprule
            \ Number of Gaussians & rFID$\downarrow$ & PSNR$\uparrow$  & SSIM$\uparrow$ \\
            \midrule
            1  & 13.11 & 20.08 & 0.57 \\
            2  & 11.08 & 20.22 & 0.59 \\
            3  & 9.10 & \textbf{20.61} & \textbf{0.61} \\
            4  & \textbf{8.38} & 20.59 & \textbf{0.61} \\
            \bottomrule
        \end{tabular}
    }
    \captionof{table}{\textbf{Ablation of Gaussian Numbers.} Increasing number of 2D Gaussians per token improves quality.}
    \label{tab:num_gs}
\end{minipage}
\end{figure}

% \begin{table*}[t]
\begin{table*}[t]
    % \centering
    \footnotesize
    \setlength{\tabcolsep}{4pt}
    \renewcommand\arraystretch{1.1}
    % 使用 minipage + \centering 强制居中
    % \begin{minipage}{\textwidth}
    \resizebox{\textwidth}{!}{ % 改为 \textwidth 更稳定
        \begin{tabular}{lcccccccccc}
        \toprule
        \multirow{2}{*}{\textbf{Method}} & \textbf{Token} & \multirow{2}{*}{\textbf{Tokens}} & \multirow{2}{*}{\textbf{Ratio}} & \textbf{Train} & \textbf{Codebook} & \textbf{Codebook} & \multirow{2}{*}{\textbf{rFID}$\downarrow$} & \multirow{2}{*}{\textbf{PSNR}$\uparrow$} & \multirow{2}{*}{\textbf{LPIPS}$\downarrow$} & \textbf{Codebook} \\ 
        & \textbf{Type} &  & & \textbf{Resolution} & \textbf{Size} & \textbf{Dim}& & & & \textbf{Usage}$\uparrow$ \\
        \midrule
        
        Titok-L~\cite{titok} & 1D & 32 & $-$ & $256 \times 256$ & 4,096 & 16 & 2.21 & $-$ & $-$  & $-$ \\
        Titok-B~\cite{titok} & 1D & 64 & $-$ & $256 \times 256$ & 4,096 & 16& 1.70 & $-$ & $-$  & $-$\\
        Titok-S~\cite{titok} & 1D & 128 & $-$ & $256 \times 256$ & 4,096 & 16 & 1.71 & $-$ & $-$ & $-$ \\
        \hline

        VQGAN~\cite{vqgan} & 2D & 16 $\times$ 16 & 16 & 256 $\times$ 256 & 1,024  & 256 & 7.94 & $-$ & $-$ & 44\% \\
        VQGAN~\cite{vqgan} & 2D & 16 $\times$ 16 & 16 & 256 $\times$ 256 & 16,384  & 256 & 4.98 & $-$ & 0.2843 & 5.9\% \\
        % VQGAN$^{*}$~\cite{vqgan} & 2D & 16 $\times$ 16 & 16 & 256 $\times$ 256 & 16,384  & 256 & 3.98  & $-$ & 0.2873 & 5.3\% \\
        SD-VQGAN~\cite{ldm} & 2D & 16 $\times$ 16 & 16 & 256 $\times$ 256 & 16,384 & 8 & 5.15 & $-$ & $-$ & $-$ \\
        MaskGIT~\cite{maskgit} & 2D & 16 $\times$ 16 & 16 & 256 $\times$ 256 & 1,024 & 256 & 2.28 & $-$ & $-$ & $-$ \\
        LlamaGen~\cite{llamagen} & 2D & 16 $\times$ 16 & 16 & 256 $\times$ 256 & 16,384 & 256 & 9.21 & 18.32 & $-$ & 0.29$\%$ \\
        LlamaGen~\cite{llamagen} & 2D & 16 $\times$ 16 & 16 & 256 $\times$ 256 & 16,384 & 8 & 2.19 & 20.79 & 0.2281 & 97$\%$ \\
        VQGAN-LC~\cite{vqganlc} & 2D & 16 $\times$ 16 & 16 & 256 $\times$ 256 & 16,384 & 8 & 3.01 & 23.2 & 0.2358 & 99$\%$ \\
        VQGAN-LC~\cite{vqganlc} & 2D & 16 $\times$ 16 & 16 & 256 $\times$ 256 & 100,000 & 8 & 2.62 & 23.8 & 0.2212 & 99$\%$ \\
        % MaskBit~\cite{maskbit} & 2D & 16 $\times$ 16 & 16 & 256 $\times$ 256 & 16,384 & 0 & 1.61 & $-$ & $-$ \\
        Open-MAGVIT2~\cite{open-magvit2} & 2D & 16 $\times$ 16 & 16 & 256 $\times$ 256 & 16,384 & 0 & 1.58 & $-$ & 0.2261 &100\% \\ 
        Open-MAGVIT2~\cite{open-magvit2} & 2D & 16 $\times$ 16 & 16 &256 $\times$ 256 & 262,144 & 0 & 1.17 & 22.64 & 0.2038  & 100\% \\
        IBQ~\cite{ibq} & 2D & 16 $\times$ 16 & 16 & 256 $\times$ 256 & 16,384 & 256 & 1.37 & $-$  & 0.2235 & 96$\%$ \\
        IBQ~\cite{ibq} & 2D & 16 $\times$ 16 & 16 & 256 $\times$ 256 & 262,144 & 256 & 1.00 & $-$ & 0.2030 & 84\% \\

        \hline
        
        \textbf{GQT} & 2D & 16 $\times$ 16 & 16 &256 $\times$ 256 & 16384 & 8 & 1.00 &22.91 & 0.1639 & 100\% \\
        \textbf{GQT-multigs} & 2D & 16 $\times$ 16 & 16 &256 $\times$ 256 & 16384 & 8 & \textbf{0.556}  & \textbf{24.93} & \textbf{0.1148} & 100\% \\        \bottomrule
        \end{tabular}
    }
    \caption{\textbf{Reconstruction performance of different tokenizers on $\boldsymbol{256 \times 256}$ ImageNet 50k validation set.} The GQT baseline already surpasses other methods, while GQT-multigs achieves a greater improvement, further validating the significant information gain introduced by the 2D Gaussians in the tokenizer architecture.}
    \label{tab:main}
    % \end{minipage}
\end{table*}

\subsection{Scaling to ImageNet}

Encouraged by the performance on Mini-ImageNet, we scale up training to the full ImageNet dataset using the same architecture and token count (256 tokens per image). All training was conducted on the NVIDIA RTX 4090 for 30 epochs to ensure a fair comparison with other methods.
Table~\ref{tab:main} summarizes the results across 15 baselines.
The original \name{} achieves the great reconstruction quality on the ImageNet 256×256 benchmark, with an \textbf{rFID score of 1.00}. This result significantly outperforms prior methods such as VQGAN (rFID = 7.94), VQGAN-LC (rFID = 3.01), and even large-scale models like Open-MAGVIT2 (rFID = 1.17), despite using a much smaller codebook. In addition to rFID, \name{} provides competitive performance in PSNR (22.91) and achieves the lowest LPIPS (0.1639) in these models. These results validate the effectiveness of our tokenizer architecture design. According to the experimental results in Table \ref{tab:num_gs}, increasing the number of 2D Gaussians associated with each token significantly enhances the model's reconstruction capability. Therefore, we tested the configuration where each token is associated with four 2D Gaussians, referred to as \textbf{\name{}-multigs}. The experimental results demonstrate that \name{}-multigs significantly outperforms both original \name{} and other methods. \name{}-multigs achieves an \textbf{rFID of 0.556 and a PSNR of 24.93, establishing a new state-of-the-art.}

All these results suggest that introducing 2D Gaussians in a dual-branch architecture allows \name{} to capture both texture and structure better, leading to more informative and compact tokenization. The outstanding performance of \name{}-multigs further reinforces the superiority of our design in reconstruction capability. Thanks to the well-optimized CUDA implementation of 2DGS, the training and inference time overhead remains minimal. Our model has a total parameter count of 180M, with 115M trainable parameters. Compared to VQ-GAN and IBQ (93.6M), the increase in parameter size is relatively modest.
In summary, our method achieves strong reconstruction performance with minimal parameter increase and a smaller codebook (compared to VQGAN-LC or Open-MAGVITV2), effectively demonstrating the advantages of integrating 2DGS into the tokenizer architecture.

\subsection{Visualization}

\begin{figure*}[t]
    \centering
    \includegraphics[width=\textwidth]{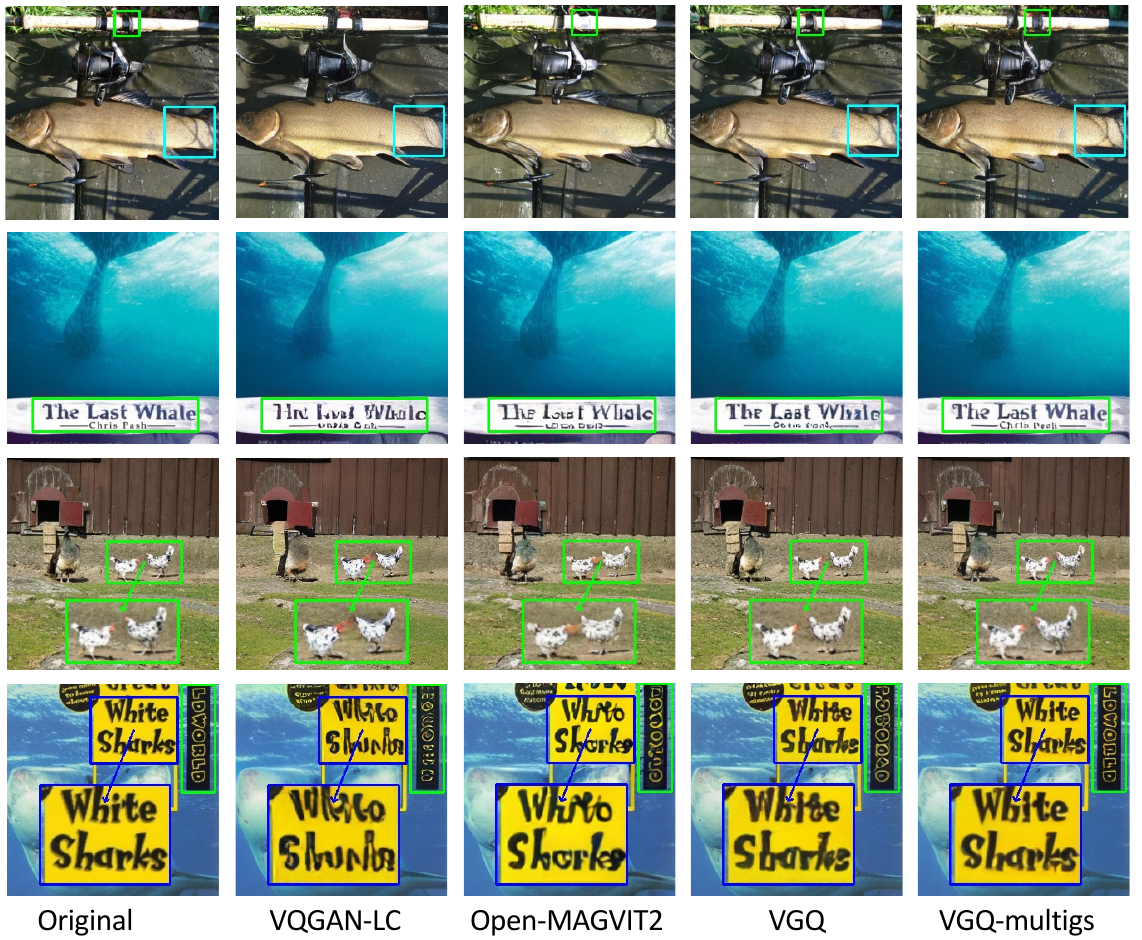}
    \caption{\textbf{Visualization Results.} Visualization results comparing reconstruction quality among VQGAN-LC, Open-MAGVIT2, \name{}, and \name{}-multigs on 256 × 256 ImageNet samples.}
    \label{fig:vis}
\end{figure*}

In this section, we present visual comparisons of reconstructed images from \name{} and \name{}-multigs against Open-MAGVITv2 and VQGAN-LC. As previously discussed, tokenizers often struggle with images containing text or intricate structures. Thus, we selected several challenging samples to illustrate these cases. Figure~\ref{fig:vis} shows visual comparisons between ground truth, images reconstructed by VQGAN-LC, Open-MAGVIT2, \name{}, and \name{}-multigs. The visualization results clearly indicate that \name{} and \name{}-multigs outperform these previous methods. The superiority is particularly evident in text regions, where \name{}-multigs maintains sharper contours and more accurate character structures. These findings further prove the effectiveness of incorporating 2D Gaussians in enhancing tokenization quality with improved structural awareness.

\section{Limitations and Future Work}
The current implementation of VGQ-multigs uses a fixed number of 2D Gaussians per token. This design may not fully take advantage of the varying complexity of different regions of the image. Future work could explore adaptive Gaussian density allocation, where regions with higher structural complexity (e.g., text or edges) are assigned more Gaussians dynamically.
VGQ is designed for image tokenization, but its potential in multimodal tasks, such as image-text generation, has not been explored. Future work could investigate how VGQ integrates with large language models or other modalities to enhance cross-modal generation and understanding.

\section{Conclusion}
We propose a novel tokenizer paradigm called Visual Gaussian Quantization (\name{}) and further extend it to \name{}-multigs. This method leverages fully discrete 2D Gaussians to tokenize images, demonstrating exceptional reconstruction capability, particularly in scenarios involving text-heavy images where traditional VQ-based tokenizers struggle. The core advantage lies in our explicit modeling of spatial attributes such as position and scales through learnable elliptical parameters. By increasing the density of 2D GS within each token, we further enhance the tokenizer’s reconstruction capability, achieving \textbf{state-of-the-art perfomance with an rFID of 0.556}.
% \section{Acknowledgments}

\bibliography{aaai2026}

\begin{thebibliography}{34}
\providecommand{\natexlab}[1]{#1}

\bibitem[{Aujol et~al.(2006)Aujol, Gilboa, Chan, and Osher}]{img-decomposition}
Aujol, J.; Gilboa, G.; Chan, T.~F.; and Osher, S.~J. 2006.
\newblock Structure-Texture Image Decomposition - Modeling, Algorithms, and Parameter Selection.
\newblock \emph{Int. J. Comput. Vis.}, 67(1): 111--136.

\bibitem[{Bachmann et~al.(2025)Bachmann, Allardice, Mizrahi, Fini, Kar, Amirloo, El-Nouby, Zamir, and Dehghan}]{flextok}
Bachmann, R.; Allardice, J.; Mizrahi, D.; Fini, E.; Kar, O.~F.; Amirloo, E.; El-Nouby, A.; Zamir, A.; and Dehghan, A. 2025.
\newblock FlexTok: Resampling Images into 1D Token Sequences of Flexible Length.
\newblock arXiv:2502.13967.

\bibitem[{Brown et~al.(2020)Brown, Mann, Ryder, Subbiah, Kaplan, Dhariwal, Neelakantan, Shyam, Sastry, Askell, Agarwal, Herbert-Voss, Krueger, Henighan, Child, Ramesh, Ziegler, Wu, Winter, Hesse, Chen, Sigler, Litwin, Gray, Chess, Clark, Berner, McCandlish, Radford, Sutskever, and Amodei}]{gpt3}
Brown, T.; Mann, B.; Ryder, N.; Subbiah, M.; Kaplan, J.~D.; Dhariwal, P.; Neelakantan, A.; Shyam, P.; Sastry, G.; Askell, A.; Agarwal, S.; Herbert-Voss, A.; Krueger, G.; Henighan, T.; Child, R.; Ramesh, A.; Ziegler, D.; Wu, J.; Winter, C.; Hesse, C.; Chen, M.; Sigler, E.; Litwin, M.; Gray, S.; Chess, B.; Clark, J.; Berner, C.; McCandlish, S.; Radford, A.; Sutskever, I.; and Amodei, D. 2020.
\newblock Language Models are Few-Shot Learners.
\newblock In \emph{NeurIPS}, volume~33, 1877--1901.

\bibitem[{Chang et~al.(2022)Chang, Zhang, Jiang, Liu, and Freeman}]{maskgit}
Chang, H.; Zhang, H.; Jiang, L.; Liu, C.; and Freeman, W.~T. 2022.
\newblock MaskGIT: Masked Generative Image Transformer.
\newblock In \emph{CVPR}, 11305--11315.

\bibitem[{Chen et~al.(2025)Chen, Li, Cai, Jiang, Qian, Kang, Gao, Zhao, Mao, and Zhang}]{halfgs}
Chen, J.; Li, Z.; Cai, Y.; Jiang, H.; Qian, C.; Kang, J.; Gao, S.; Zhao, H.; Mao, T.; and Zhang, Y. 2025.
\newblock {HAIF-GS:} Hierarchical and Induced Flow-Guided Gaussian Splatting for Dynamic Scene.
\newblock \emph{CoRR}, abs/2506.09518.

\bibitem[{Chowdhery et~al.(2023)Chowdhery, Narang, Devlin, Bosma, Mishra, Roberts, Barham, Chung, Sutton, Gehrmann, Schuh, Shi, Tsvyashchenko, Maynez, Rao, Barnes, Tay, Shazeer, Prabhakaran, Reif, Du, Hutchinson, Pope, Bradbury, Austin, Isard, Gur{-}Ari, Yin, Duke, Levskaya, Ghemawat, Dev, Michalewski, Garcia, Misra, Robinson, Fedus, Zhou, Ippolito, Luan, Lim, Zoph, Spiridonov, Sepassi, Dohan, Agrawal, Omernick, Dai, Pillai, Pellat, Lewkowycz, Moreira, Child, Polozov, Lee, Zhou, Wang, Saeta, Diaz, Firat, Catasta, Wei, Meier{-}Hellstern, Eck, Dean, Petrov, and Fiedel}]{palm}
Chowdhery, A.; Narang, S.; Devlin, J.; Bosma, M.; Mishra, G.; Roberts, A.; Barham, P.; Chung, H.~W.; Sutton, C.; Gehrmann, S.; Schuh, P.; Shi, K.; Tsvyashchenko, S.; Maynez, J.; Rao, A.; Barnes, P.; Tay, Y.; Shazeer, N.; Prabhakaran, V.; Reif, E.; Du, N.; Hutchinson, B.; Pope, R.; Bradbury, J.; Austin, J.; Isard, M.; Gur{-}Ari, G.; Yin, P.; Duke, T.; Levskaya, A.; Ghemawat, S.; Dev, S.; Michalewski, H.; Garcia, X.; Misra, V.; Robinson, K.; Fedus, L.; Zhou, D.; Ippolito, D.; Luan, D.; Lim, H.; Zoph, B.; Spiridonov, A.; Sepassi, R.; Dohan, D.; Agrawal, S.; Omernick, M.; Dai, A.~M.; Pillai, T.~S.; Pellat, M.; Lewkowycz, A.; Moreira, E.; Child, R.; Polozov, O.; Lee, K.; Zhou, Z.; Wang, X.; Saeta, B.; Diaz, M.; Firat, O.; Catasta, M.; Wei, J.; Meier{-}Hellstern, K.; Eck, D.; Dean, J.; Petrov, S.; and Fiedel, N. 2023.
\newblock PaLM: Scaling Language Modeling with Pathways.
\newblock \emph{J. Mach. Learn. Res.}, 24: 240:1--240:113.

\bibitem[{Dong et~al.(2025)Dong, Wang, Zheng, Chen, Lu, and Tang}]{gstk}
Dong, J.; Wang, C.; Zheng, W.; Chen, L.; Lu, J.; and Tang, Y. 2025.
\newblock GaussianToken: An Effective Image Tokenizer with 2D Gaussian Splatting.
\newblock arXiv:2501.15619.

\bibitem[{Esser, Rombach, and Ommer(2021)}]{vqgan}
Esser, P.; Rombach, R.; and Ommer, B. 2021.
\newblock Taming Transformers for High-Resolution Image Synthesis.
\newblock In \emph{CVPR}, 12873--12883.

\bibitem[{Huang et~al.(2024)Huang, Yu, Chen, Geiger, and Gao}]{2dgs}
Huang, B.; Yu, Z.; Chen, A.; Geiger, A.; and Gao, S. 2024.
\newblock 2D Gaussian Splatting for Geometrically Accurate Radiance Fields.
\newblock In Burbano, A.; Zorin, D.; and Jarosz, W., eds., \emph{{ACM} {SIGGRAPH} 2024 Conference Papers, {SIGGRAPH} 2024, Denver, CO, USA, 27 July 2024- 1 August 2024}, 32. {ACM}.

\bibitem[{Huang et~al.(2023)Huang, Mao, Chen, and Zhang}]{dqvae}
Huang, M.; Mao, Z.; Chen, Z.; and Zhang, Y. 2023.
\newblock Towards accurate image coding: Improved autoregressive image generation with dynamic vector quantization.
\newblock In \emph{CVPR}, 22596--22605.

\bibitem[{Isola et~al.(2017)Isola, Zhu, Zhou, and Efros}]{patchgan}
Isola, P.; Zhu, J.; Zhou, T.; and Efros, A.~A. 2017.
\newblock Image-to-Image Translation with Conditional Adversarial Networks.
\newblock In \emph{CVPR}, 5967--5976.

\bibitem[{Kerbl et~al.(2023)Kerbl, Kopanas, Leimk{\"{u}}hler, and Drettakis}]{3dgs}
Kerbl, B.; Kopanas, G.; Leimk{\"{u}}hler, T.; and Drettakis, G. 2023.
\newblock 3D Gaussian Splatting for Real-Time Radiance Field Rendering.
\newblock \emph{{ACM} Trans. Graph.}, 42(4): 139:1--139:14.

\bibitem[{Lee et~al.(2022)Lee, Kim, Kim, Cho, and Han}]{rqvae}
Lee, D.; Kim, C.; Kim, S.; Cho, M.; and Han, W. 2022.
\newblock Autoregressive Image Generation using Residual Quantization.
\newblock In \emph{CVPR}, 11513--11522.

\bibitem[{Liu et~al.(2024)Liu, Li, Wu, and Lee}]{llava}
Liu, H.; Li, C.; Wu, Q.; and Lee, Y.~J. 2024.
\newblock Visual instruction tuning.
\newblock \emph{Advances in neural information processing systems}, 36.

\bibitem[{Lorenz et~al.(2019)Lorenz, Bereska, Milbich, and Ommer}]{disentangle-obj-appearance}
Lorenz, D.; Bereska, L.; Milbich, T.; and Ommer, B. 2019.
\newblock Unsupervised Part-Based Disentangling of Object Shape and Appearance.
\newblock In \emph{{IEEE} Conference on Computer Vision and Pattern Recognition, {CVPR} 2019, Long Beach, CA, USA, June 16-20, 2019}, 10955--10964. Computer Vision Foundation / {IEEE}.

\bibitem[{Luo et~al.(2024)Luo, Shi, Ge, Yang, Wang, and Shan}]{open-magvit2}
Luo, Z.; Shi, F.; Ge, Y.; Yang, Y.; Wang, L.; and Shan, Y. 2024.
\newblock Open-magvit2: An open-source project toward democratizing auto-regressive visual generation.
\newblock \emph{arXiv preprint arXiv:2409.04410}.

\bibitem[{Mentzer et~al.()Mentzer, Minnen, Agustsson, and Tschannen}]{fsq}
Mentzer, F.; Minnen, D.; Agustsson, E.; and Tschannen, M. ????
\newblock Finite Scalar Quantization: {VQ-VAE} Made Simple.
\newblock In \emph{The Twelfth International Conference on Learning Representations, {ICLR} 2024, Vienna, Austria, May 7-11, 2024}.

\bibitem[{OpenAI(2023)}]{openai2023gpt4}
OpenAI. 2023.
\newblock {GPT}-4 technical report.
\newblock \emph{arXiv preprint arXiv:2303.08774}.

\bibitem[{Qu et~al.(2024)Qu, Zhang, Liu, Wang, Jiang, Gao, Ye, Du, Yuan, and Wu}]{tokenflow}
Qu, L.; Zhang, H.; Liu, Y.; Wang, X.; Jiang, Y.; Gao, Y.; Ye, H.; Du, D.~K.; Yuan, Z.; and Wu, X. 2024.
\newblock TokenFlow: Unified Image Tokenizer for Multimodal Understanding and Generation.
\newblock \emph{arXiv preprint arXiv:2412.03069}.

\bibitem[{Rombach et~al.(2022)Rombach, Blattmann, Lorenz, Esser, and Ommer}]{ldm}
Rombach, R.; Blattmann, A.; Lorenz, D.; Esser, P.; and Ommer, B. 2022.
\newblock High-resolution image synthesis with latent diffusion models.
\newblock In \emph{CVPR}, 10684--10695.

\bibitem[{Sanchez, Serrurier, and Ortner(2020)}]{mutual-information}
Sanchez, E.~H.; Serrurier, M.; and Ortner, M. 2020.
\newblock Learning Disentangled Representations via Mutual Information Estimation.
\newblock In Vedaldi, A.; Bischof, H.; Brox, T.; and Frahm, J., eds., \emph{Computer Vision - {ECCV} 2020 - 16th European Conference, Glasgow, UK, August 23-28, 2020, Proceedings, Part {XXII}}, volume 12367 of \emph{Lecture Notes in Computer Science}, 205--221. Springer.

\bibitem[{Shi et~al.(2025)Shi, Luo, Ge, Yang, Shan, and Wang}]{ibq}
Shi, F.; Luo, Z.; Ge, Y.; Yang, Y.; Shan, Y.; and Wang, L. 2025.
\newblock Scalable Image Tokenization with Index Backpropagation Quantization.
\newblock arXiv:2412.02692.

\bibitem[{Simonyan and Zisserman(2015)}]{vgg}
Simonyan, K.; and Zisserman, A. 2015.
\newblock Very Deep Convolutional Networks for Large-Scale Image Recognition.
\newblock In Bengio, Y.; and LeCun, Y., eds., \emph{3rd International Conference on Learning Representations, {ICLR} 2015, San Diego, CA, USA, May 7-9, 2015, Conference Track Proceedings}.

\bibitem[{Sun et~al.(2024)Sun, Jiang, Chen, Zhang, Peng, Luo, and Yuan}]{llamagen}
Sun, P.; Jiang, Y.; Chen, S.; Zhang, S.; Peng, B.; Luo, P.; and Yuan, Z. 2024.
\newblock Autoregressive Model Beats Diffusion: Llama for Scalable Image Generation.
\newblock \emph{arXiv preprint arXiv:2406.06525}.

\bibitem[{Touvron et~al.(2023)Touvron, Martin, Stone, Albert, Almahairi, Babaei, Bashlykov, Batra, Bhargava, Bhosale, Bikel, Blecher, Canton{-}Ferrer, Chen, Cucurull, Esiobu, Fernandes, Fu, Fu, Fuller, Gao, Goswami, Goyal, Hartshorn, Hosseini, Hou, Inan, Kardas, Kerkez, Khabsa, Kloumann, Korenev, Koura, Lachaux, Lavril, Lee, Liskovich, Lu, Mao, Martinet, Mihaylov, Mishra, Molybog, Nie, Poulton, Reizenstein, Rungta, Saladi, Schelten, Silva, Smith, Subramanian, Tan, Tang, Taylor, Williams, Kuan, Xu, Yan, Zarov, Zhang, Fan, Kambadur, Narang, Rodriguez, Stojnic, Edunov, and Scialom}]{llama2}
Touvron, H.; Martin, L.; Stone, K.; Albert, P.; Almahairi, A.; Babaei, Y.; Bashlykov, N.; Batra, S.; Bhargava, P.; Bhosale, S.; Bikel, D.; Blecher, L.; Canton{-}Ferrer, C.; Chen, M.; Cucurull, G.; Esiobu, D.; Fernandes, J.; Fu, J.; Fu, W.; Fuller, B.; Gao, C.; Goswami, V.; Goyal, N.; Hartshorn, A.; Hosseini, S.; Hou, R.; Inan, H.; Kardas, M.; Kerkez, V.; Khabsa, M.; Kloumann, I.; Korenev, A.; Koura, P.~S.; Lachaux, M.; Lavril, T.; Lee, J.; Liskovich, D.; Lu, Y.; Mao, Y.; Martinet, X.; Mihaylov, T.; Mishra, P.; Molybog, I.; Nie, Y.; Poulton, A.; Reizenstein, J.; Rungta, R.; Saladi, K.; Schelten, A.; Silva, R.; Smith, E.~M.; Subramanian, R.; Tan, X.~E.; Tang, B.; Taylor, R.; Williams, A.; Kuan, J.~X.; Xu, P.; Yan, Z.; Zarov, I.; Zhang, Y.; Fan, A.; Kambadur, M.; Narang, S.; Rodriguez, A.; Stojnic, R.; Edunov, S.; and Scialom, T. 2023.
\newblock Llama 2: Open Foundation and Fine-Tuned Chat Models.
\newblock \emph{arXiv preprint arXiv:2307.09288}.

\bibitem[{Van Den~Oord, Vinyals et~al.(2017)}]{vqvae}
Van Den~Oord, A.; Vinyals, O.; et~al. 2017.
\newblock Neural discrete representation learning.
\newblock In \emph{NeurIPS}, volume~30.

\bibitem[{Wu et~al.(2024)Wu, Yi, Fang, Xie, Zhang, Wei, Liu, Tian, and Wang}]{4dgs}
Wu, G.; Yi, T.; Fang, J.; Xie, L.; Zhang, X.; Wei, W.; Liu, W.; Tian, Q.; and Wang, X. 2024.
\newblock 4D Gaussian Splatting for Real-Time Dynamic Scene Rendering.
\newblock In \emph{{IEEE/CVF} Conference on Computer Vision and Pattern Recognition, {CVPR} 2024, Seattle, WA, USA, June 16-22, 2024}, 20310--20320. {IEEE}.

\bibitem[{Yu et~al.(2023)Yu, Cheng, Sohn, Lezama, Zhang, Chang, Hauptmann, Yang, Hao, Essa, and Jiang}]{magvit1}
Yu, L.; Cheng, Y.; Sohn, K.; Lezama, J.; Zhang, H.; Chang, H.; Hauptmann, A.~G.; Yang, M.; Hao, Y.; Essa, I.; and Jiang, L. 2023.
\newblock {MAGVIT:} Masked Generative Video Transformer.
\newblock In \emph{CVPR}, 10459--10469.

\bibitem[{Yu et~al.(2024{\natexlab{a}})Yu, Lezama, Gundavarapu, Versari, Sohn, Minnen, Cheng, Gupta, Gu, Hauptmann, Gong, Yang, Essa, Ross, and Jiang}]{magvit2}
Yu, L.; Lezama, J.; Gundavarapu, N.~B.; Versari, L.; Sohn, K.; Minnen, D.; Cheng, Y.; Gupta, A.; Gu, X.; Hauptmann, A.~G.; Gong, B.; Yang, M.-H.; Essa, I.; Ross, D.~A.; and Jiang, L. 2024{\natexlab{a}}.
\newblock Language Model Beats Diffusion - Tokenizer is key to visual generation.
\newblock In \emph{ICLR}.

\bibitem[{Yu et~al.(2024{\natexlab{b}})Yu, Weber, Deng, Shen, Cremers, and Chen}]{titok}
Yu, Q.; Weber, M.; Deng, X.; Shen, X.; Cremers, D.; and Chen, L.-C. 2024{\natexlab{b}}.
\newblock An Image is Worth 32 Tokens for Reconstruction and Generation.
\newblock \emph{arXiv preprint arXiv:2406.07550}.

\bibitem[{Zhang et~al.(2024{\natexlab{a}})Zhang, Ge, Xu, He, Wang, Qin, Lu, Geng, and Zhang}]{gsimg}
Zhang, X.; Ge, X.; Xu, T.; He, D.; Wang, Y.; Qin, H.; Lu, G.; Geng, J.; and Zhang, J. 2024{\natexlab{a}}.
\newblock GaussianImage: 1000 FPS Image Representation and Compression by 2D Gaussian Splatting.
\newblock arXiv:2403.08551.

\bibitem[{Zhang et~al.(2024{\natexlab{b}})Zhang, Kuznetsov, Jindal, Chen, Sochenov, Kaplanyan, and Sun}]{image-gs}
Zhang, Y.; Kuznetsov, A.; Jindal, A.; Chen, K.; Sochenov, A.; Kaplanyan, A.; and Sun, Q. 2024{\natexlab{b}}.
\newblock Image-GS: Content-Adaptive Image Representation via 2D Gaussians.
\newblock \emph{CoRR}, abs/2407.01866.

\bibitem[{Zhu et~al.(2023)Zhu, Chen, Shen, Li, and Elhoseiny}]{minigpt}
Zhu, D.; Chen, J.; Shen, X.; Li, X.; and Elhoseiny, M. 2023.
\newblock Minigpt-4: Enhancing vision-language understanding with advanced large language models.
\newblock \emph{arXiv preprint arXiv:2304.10592}.

\bibitem[{Zhu et~al.(2024)Zhu, Wei, Lu, and Chen}]{vqganlc}
Zhu, L.; Wei, F.; Lu, Y.; and Chen, D. 2024.
\newblock Scaling the Codebook Size of VQGAN to 100,000 with a Utilization Rate of 99\%.
\newblock \emph{arXiv preprint arXiv:2406.11837}.

\end{thebibliography}

\end{document}